\documentclass[journal]{IEEEtran}

 \usepackage{graphicx}
 \usepackage{url}
\usepackage{hyperref}
\usepackage[utf8]{inputenc}
\usepackage{color, xcolor}
\usepackage{color}

\begin{document}
\title{Explainable Artificial Intelligence (XAI):  An Engineering Perspective}

\author{
	Fatima Hussain,
	Rasheed Hussain, {\em SMIEEE}, and
	Ekram Hossain, {\em FIEEE}
	
	\thanks{
		
F. Hussain is with User Behaviour Analytics and Insider Threat Team, Royal Bank of Canada, Toronto, Canada (email: fatima.hussain@rbc.com).
		
		R. Hussain is with Networks and Blockchain Laboratory, Innopolis University, Innopolis, Russia (email: r.hussain@innopolis.ru).
		
		E. Hossain is with Department of Electrical and Computer Engineering at University of Manitoba, Winnipeg, Canada (email: Ekram.Hossain@umanitoba.ca). His work was supported by a Discovery Grant from the Natural Sciences and Engineering Research Council of Canada (NSERC).
	}
}

\maketitle

\begin{abstract}
The remarkable advancements in Deep Learning (DL) algorithms have fueled enthusiasm for using Artificial Intelligence (AI) technologies in almost every domain; however, the opaqueness of these algorithms put a question mark on their applications in safety-critical systems. In this regard, the `explainability' dimension is not only essential to both explain the inner workings of black-box algorithms, but it also adds accountability and transparency dimensions that are of prime importance for regulators, consumers, and service providers. eXplainable Artificial Intelligence (XAI) is the set of techniques and methods to convert the so-called black-box AI algorithms to white-box algorithms, where the results achieved by these algorithms and the variables, parameters, and steps taken by the algorithm to reach the obtained results, are transparent and explainable. To complement the existing literature on XAI, in this paper, we take an `engineering’ approach to illustrate the concepts of XAI. We discuss the stakeholders in XAI and describe the mathematical contours of XAI from engineering perspective. Then we take the autonomous car as a use-case and discuss the applications of XAI for its different components such as object detection, perception, control, action decision, and so on. This work is an exploratory study to identify new avenues of research in the field of XAI.  
 \end{abstract}


\section{Introduction and Basic Concepts}
 
% \cite{Wang}, \cite{Anjomshoae}, \cite{Wolf}, \cite{Amina}

\subsection{Background}
The inception of futuristic technologies such as the Internet of Things (IoT), autonomous driving, augmented, and virtual reality, at least in part, owe to the advancements in communication technologies. Similarly, the idea of smart `{\it things}' has revolutionized the applications and services space which directly affects our lives in a positive way. %Today’s disruptive technologies are inherently and massively heterogeneous covering different domains, for instance, but not limited to, smart home, smart industry, health, business, retail, and so on. However, there is a common phenomenon in the applications realized through the aforementioned technologies, that is, a huge amount of data with varying speed, velocity, veracity, and volume is produced by the entities participating in realizing these technologies. 
These applications generate huge amount of high dimensional and heterogeneous data, and  efficient methods are required to extract valuable information from such  data. 
%%% Un-commented by Rasheed on Dec 7, 2020
For instance, an autonomous car is expected to produce 5 to 20 Terabytes of data every day (depending on the duration of driving)\footnote{\protect \url{https://cutt.ly/nh8rMyj}}. These data may require real-time or near-real-time processing for monitoring, prediction, decision making, and control purposes (e.g., for autonomous navigation). Artificial Intelligence (AI) and its breeds, Machine Learning (ML), and Deep Learning (DL) techniques are primarily used as data analytics tools in such scenarios. 

To date, remarkable research results have been achieved by using ML and DL in futuristic technologies. For instance, ML and DL are widely used in autonomous cars for lane and object detection, and perception, mapping, planning, route calculation, actuation, and so on \cite{Hussain2018,Grigorescu2020,Ma2020}. 
%Un-commented by Rasheed on December 7, 2020
Furthermore,  cloud and fog computing are becoming key components for many technologies and applications. Data-driven machine learning will be a key ingredient in cloud and fog computing platforms.  

%Talk about blackbox idea in ML and DL.
Despite the success of ML- and DL-based solutions in different domains, there are a number of challenges faced by ML and DL. For instance, the overhead incurred by the ML- and DL-based solutions in different domains such as IoT networks, healthcare, autonomous driving, and so on, will greatly affect their scalability. Furthermore, outsourcing the data generated by the users in every domain, may infringe the privacy whereas the management of such data is another challenge. %{\todo (need for transparency)} 

It is worth mentioning that depending on the domain, the use of ML and DL is extremely critical from the point of view of decisions taken by these algorithms. In such cases, even a slight dysfunction of an ML/DL algorithm can have catastrophic consequences. For instance, in case of an autonomous car, even a minor problem in computer vision component may lead to fatality. A similar situation may occur in other domains such as healthcare during diagnosis, prognosis, and surgery. The {\em explainability}, {\em transparency}, and  {\em accountability} features are generally missing in the existing ML and DL models. For instance, if there is a situation where the ML/DL model's output is undesirable (e.g., an autonomous car takes an unexpected decision), then the accountability factor becomes of paramount importance. Who should be accountable for that decision? Why did the ML/DL model came to that conclusion? These questions need to be answered. 

%In essence, the decisions taken on the basis of ML and explicitly DL models, are not inherently explainable. More precisely, non-linear models such as DL models exhibit more complexity than the linear ML models in terms of explanation. 

% where the outputs of the model are heavily dependent on the inputs and changing inputs may have critical impact on other inputs and thus outputs as well. 
%%%% Added by Rasheed on January 1, 2021 in response to Prof. E. Hossain's comments

\subsection{Basic Attributes of XAI}

When considering the comprehensiveness of the AI models \cite{Arrieta}, there are three dimensions as discussed below.
\begin{enumerate}
    \item {\it Explainability}: It is an active feature of a learning model through which the processes undertaken by the model can be clearly described. The aim is to clarify the inner working of the learning model. Note that, critical applications need explanability not just for intellectual curiosity, but the risk factor is weighed above all other factors when human lives are in danger \cite{Xie2019}.
    \item {\it Interpretability}: Unlike explainability, it is a passive feature of a learning model which enables the users to understand the model and make sense out of it.
    \item {\it Transparency}: Transparency is also directly related to understandability where a learning model is considered transparent if it exhibits understandability on its own and without any interface. When a learning model is inherently  understandable without any extra components introduced to the model, the model is transparent. 
\end{enumerate}

From the above definitions, it is also clear that transparency embodies the explainability and interpretability and hence emerges as the strong dominating aspect of comprehensiveness in a learning model. Interpretability in ML and DL models has been a hot topic of debate in the academia and industry \cite{Tjoa2020}. According to Tjoa et al., interpretability has inherent benefits of improving the performance of ML and DL models in terms of causality, reliability, and usability \cite{Tjoa2020}. We also note that different literature provide different definitions for explainability and interpretability (as discussed above) \cite{das2020,Amina2018,Amina2019,Arrieta,Samek2019}; however, Tjoa et al. used these two terms interchangeably. According to their definition,  interpretability (and thus explainability) is a mechanism by which the decisions taken by the algorithms can be explained,  the inner details of the model are understandable, and the models can be explained mathematically. In our discussion, without loss of generality, we will follow the approach of Tjoa et al. and use the terms interpretability and explanability interchangeably. 

The explainability feature is essential for the AI models\footnote{While discussing XAI, we focus on the breeds of AI, i.e., ML and DL models. Therefore, in this paper, although we mention ML and DL models, in fact are the AI models.}, and their decisions, to be transparent. 
On the other hand, the recent threats and adversarial attacks on ML and DL algorithms also warrant the need for algorithmic and functional transparency in ML and DL mechanisms \cite{Akhtar2018,MA2021,REN2020}. Without loss of generality, the transparency in AI models will serve several-fold purposes, for instance, it will increase the trust in AI with positive expected outcomes, enable the decision to whether fully rely on AI or consider human-factor for decision-making, and address the security attacks on AI-based techniques. 

Another important dimension of XAI is {\em accountability}, which is a legal dimension. Amidst the implementation of General Data Protection Regulation (GDPR), the legal dimension of accountability becomes more prevalent because of the liability issues in the emerging applications that deal with privacy-sensitive data (e.g., autonomous cars, IoT, augmented reality, etc.). In such situations (to be compliant with GDPR), risk assessment can be carried out to decrease the liability issues. For instance, in case of healthcare IoT applications that collect and aggregate data from Body Area Network (BAN) and then the data is either locally processed in a mobile device or sent to a cloud storage for further processing. The results obtained from such data could be shared with different stakeholders including doctors, insurance agencies, and so on. The risk of sharing this data with different entities (with proper access control mechanism) should also take into account the liability issues in case of any mishap.

In the light of the above discussion, a new term Explainable Artificial Intelligence (XAI) was tossed that adds the explainability, transparency, and accountability dimensions to the existing ML and DL (and thus AI) models \cite{Amina2018}. In other words, XAI is an effort to convert the black-box AI models to white-box approaches where the overall process taken by the algorithms and models to reach a decision can be {\it explained and interpreted,} categorically. 
XAI is a Defense Advanced Research Projects Agency (DARPA) initiative, and it enables AI systems to have the feature of a {\it glass box} in which machines can understand the context in their operating environment and with human in the loop of the decisions. The other objective of XAI is to build underlying explanatory models for characterizing real world phenomena. DARPA has divided XAI into three categories namely, {\it Deep Explanation}, {\it Interpretable Models}, and  {\it Model Induction}. Broadly speaking, we need two distinct systems, one which can interpret the already existing complex models, and second which can explain the newly designed models and the decision they make \cite{Brian,Hani,Arrieta}.  

\subsection{Contribution and Organization of the Paper}

The existing literature on XAI takes the application-level approach to explain the current ML and DL algorithms and argue on transparency. On the other hand, in this paper, we take an engineering perspective of the explainability and transparency of the ML and DL mechanisms. 
More precisely, we explain the explainability from an engineering perspective by explaining the problem of AI models that need {\it explanation} and focus on the mathematical perspective of explanation. We discuss different stakeholders of XAI and their perception and requirements of XAI. Moving forward, we discuss the engineering and mathematical perspective of XAI and take autonomous car as a use-case of XAI. In the autonomous car use-case, we identify and discuss in detail, the aspects that need and benefit from explainable AI models, and discuss the current solutions for XAI in autonomous cars. More precisely, from an engineering perspective, we identify the problems of AI models used in different components of autonomous car such as object detection, control, perception, etc. and discuss the existing solutions towards adding explainability features to the AI models used in autonomous car. The aim of this paper is to spur further discussion on the need for XAI in critical applications and look at XAI in critical applications from engineering perspective.  The autonomous car use-case is selected because it covers the engineering domain where we define the clear problems pertaining to different components of autonomous.

\section{XAI Systems and Techniques} 
\label{sec:XAISystem}
In the literature, various models are developed and explained for the explanation of the existing ML models; however, only explanations about, ``how the prediction is made?,'' is not enough to justify the validity of results. This white box model is not useful due to the complexities of most of the models, which are also quantitative and non-intuitive in nature.
Therefore, users (who are not ML experts) are unable to understand the functionalities of the model, no matter how transparent the model is. %much transparency is presented. 
Hence, we require interpretable models to produce comprehensive details of the decisions and predictions made by the models.

In the following, we discuss these two broad categories of XAI systems. 

\subsection{Transparent Models} 
These models are also known as interpretable models and are interpretable by design, in contrast to black-box ML models. These transparent models refer to explanation and understandability from the perspective of algorithmic transparency followed by algorithm decomposability as well as simulation ability of a model.  All of these explanation types define various levels of explanations that a specific model is capable of providing. 

First level of transparency of any model is its algorithmic transparency; i.e., how well a user can understand the output obtained from a given set of inputs. More specifically, transparent model is fully explorable with the help of  mathematical analysis and methods. Linear models are easy to interpret whereas non-linear models require more sophisticated methods.
%, such as, heuristic optimization techniques to fully understand the deep architecture of the opaque model.  

Second level of transparency is the level of decomposability of a model. This means, how intelligent an explanation of each part (input, various parameters, and calculation performed) of a model is? As a result of this explanation, it is easier to understand and explain the behavior of a model. However, it requires all the inputs to be readily interpretable which is  not possible for every model. % to fulfill this property. 
This is due to the reason that cumbersome and complex features and parameters might not  fulfill these requirements, and cannot be understood without incorporating additional tools.

Furthermore, a completely transparent model or third level of transparency is achieved if the model can be simulated. It is the ability of the model to be simulated by humans. Complexity of any model is definitely having prominence in this transparency level.

%Therefore, sparse linear models are more interpretable than complex . Same is true for rule based systems, having large amount of rules and parameters, in contrast to perception neural networks that have only single layer and are using linear classifiers.

In conclusion, an interpretable model can be easily explained to  humans with the help of text and visualizations. Furthermore, decomposable model having ability to be simulated means that,  model is self-contained in a way that it can be understood, analyzed, and justified as a whole (by humans without need of additional tools).

\begin{figure*}
\centerline{\includegraphics[scale=0.55]{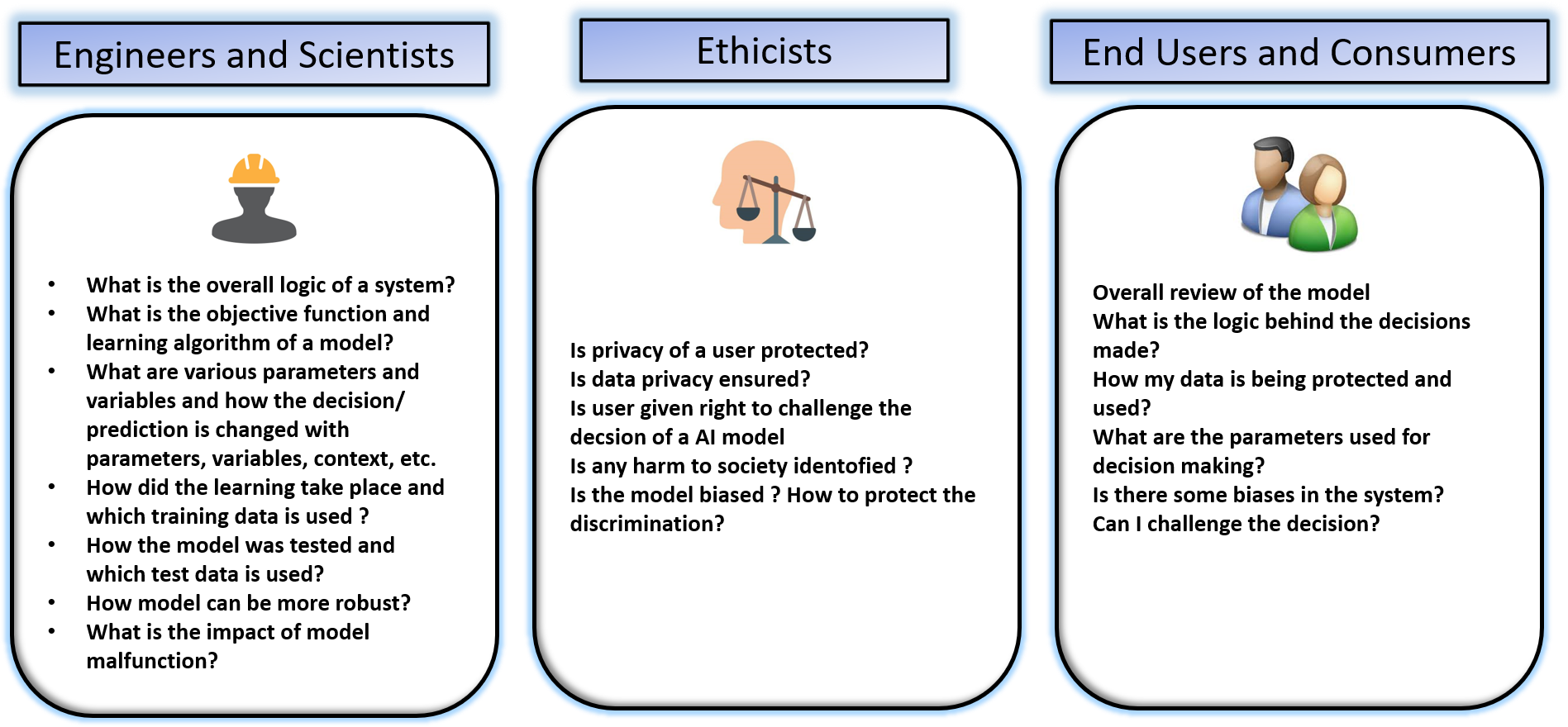}}
\caption{Stakeholders in XAI}
\label{fig:stakeholders}
\end{figure*}

\subsection{Prediction Interpretation and Justification Models} 
Black box models that are not interpretable by design, require post-hoc explainability to enhance their interpretability by various human-adopted explanation methods. These expalinability methods can be text explanations, visual explanations, explanations by simplification, explanations by example, and explanations by feature relevance. Broadly speaking, such post-hoc explainability depends upon the explanation techniques adopted by the interpreters (authors), types of data on which they applied and actual technique being used. For instance, interpreter (author) can use explanation by simplification and perform sensitivity analysis on images. Or the interpreter can either choose to perform explanation by using different symbols or by various examples, etc. 

In \cite{Doran}, the authors discuss various types of black box models and presented the a broader distinction on the basis of transparency inferred by the users. The authors classify the systems into three types: opaque, interpretable, and comprehensible  systems. In opaque systems, the mappings from input to output  is invisible to the user, while in interpretable systems, users are aware of and can mathematically analyze the input to output mappings. Lastly, in comprehensible systems, users are not only aware of the mapping but also specific rules behind those mappings. In short, comprehensible systems provide a comprehensive view of symbols (rules) of mapping  along with their specific output in a way that is completely understood by the users. 

These  post-hoc explanation techniques can be applied to infer an interpretable model from any black box model. The techniques used to infer an explainable model from any black box model are also know as model induction \cite{Hani}. However, it is difficult to interpret the entire model with comprehensive explanation, unless complete description of the model is performed. Local or global context is also an important factor to consider, i.e., explanation in local fidelity might not be completely true in global fidelity \cite{singh}.

 \section{Stakeholders in XAI and Explainability Requirements}

As discussed earlier, an intelligent system cannot be completely trusted if it does not provide enough justification for the predictions it makes. It is more trusted by clients if the system is explained as per caliber and explanation requirement of various stakeholders of that system. To this end, %We believe that 
the most important task is to identify the stakeholders and afterwards, provide acceptable explanation and interpretation for each type of recipient (such as, engineers, designers, theorists etc.).

Before we dive into scientific and mathematical perspective of XAI, in the following, we briefly discuss various stakeholders and associated requirements to construct the narrative.  

%Since, our current work in XAI focuses on the %is with 
%perspective of engineers, therefore, our discussion will surround the engineering perspective. %we will focus more in this context. 

Once the mathematical description of ML models is available, testing (functionality) the model is possible not only through mathematical analysis (for certain attributes) but can also be checked for operational validity. This is done by feeding new input data to the model and obtaining anticipated correct output, thus validating the accuracy of the model (the model under consideration). This engineering and mathematical perspective may not be required for an ordinary customer but is of utmost importance for a system designer and developer. They can understand such a system in a better way if it is more transparent and interpretable by design and at development stage.

\subsection{Stakeholders in XAI}
In this section, we discuss different stakeholders in XAI and shed light on their role as well as their expectations from XAI. 

\subsubsection{End-users and consumers}
End users and consumers need explanations to justify the actions of a system and decide
whether the outputs of the system are correct or not. In fact, assurance and confidence should be given to users through post-hoc explanation (with policy and ontology elements), that the system does the right thing and give the unbiased expected outcome. 
From the end-user perspective, explanation of an XAI system should always reflect
the psychological processes of human understanding and essentially must follow the following basic principles:
\begin{itemize}
    \item User is able to understand the conceptual model and underlying mechanisms of the system.
\item Entire functionality of a system is visible to users and enough explanations are given about the capability of the the system. % is capable of doing.
\item The visible elements of the system can be intuitively mapped to respective functionalities.
\item Users are always informed about the system's current state.
\end{itemize}

\subsubsection{Ethicists and theorists}
Ethicists are more concerned about distinct explanations regarding transparency and fairness of an AI systems, and include various stakeholders such as computer scientists, engineers, lawyers, journalists,
economists, and politicians. The sole purpose of these stakeholders is to go beyond the technical details and assure fairness and unbiased behaviour in AI system, thus making it accountable. Furthermore, assurance and confidence are required by these ethicists through post-hoc explanations that the system has reached ethical decisions without any biases. 

Theorists, on the other hand are more interested in understanding the theory behind AI and further advancing it. These stakeholders include %various stakeholders such as  
academic or industrial researchers who are not much interested in delivering the practical applications. These stakeholders also demand transparent relationship between internal states of the model and the achieved output. 

\subsubsection{Engineers and mathematicians} Mathematical models are quantitative ways to describe both theoretical and real-world systems. Engineers use mathematical models to describe the behavior of any newly designed system or examine an existing one.
Therefore, once we demand or define explanation for engineers and scientists,  it is not just the data description or high-level discussion about various decisions being made by that system. They also expect transparency-based bindings to the internal states of the model 
(i.e., traceability to any action/state is possible), such that these explanations are not just post-hoc rationalisation.   

 \begin{figure*}
%\centerline{\includegraphics[scale=0.6]{Model.png}}
% [width=8in,height=6in]
\centerline{\includegraphics[width=7in]{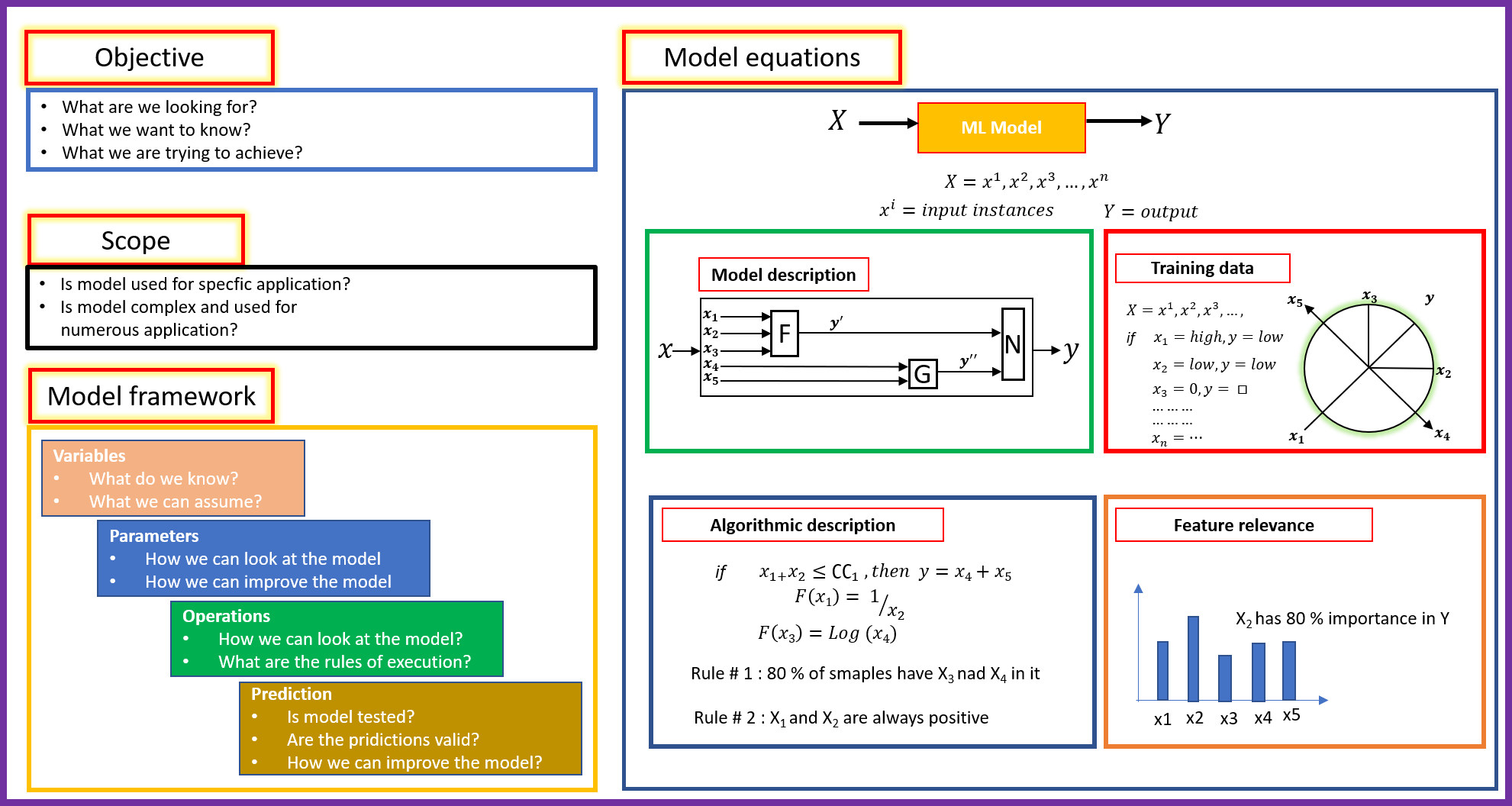}}
\caption{Mathematical perspective of XAI model}
\label{fig:Model}
\end{figure*}
\subsection{Role-Based Explainability Requirements}
%{\todo \bf (A very good paper for stakeholders (\url{https://arxiv.org/pdf/1810.00184.pdf}))}

It is  important to identify the stakeholder for explanation, i.e., whom are the explanations for? Is it for advanced mathematicians or engineers, or employees and customers? There should be complete definition of the ``Explainer'' roles to make AI models more relevant and readily interpretable. Furthermore, a clear definition of ``What'' is required to carry out the ``explanation''. It is at par important to understand the subject of explanation and to figure out whether it is the model or the data itself. %is required to understand the model or data itself? 
For instance, consider a system used for identification of various cars on road. An explanation of data (obtained from a source)  may include the description that a particular image is of a car as it is similar (in shape and dimensions) to another image of a car (KNN application), or the decision is made by considering various features (feature selection) like four wheels, shape, back and front lights, etc.  

In a nutshell, comprehensibility and explainability should be considered in the
context of a specific use-case or a stakeholder \cite{Richard,Preece2018}. To put this discussion into perspective, we discuss the use-case of autonomous car in detail in Section \ref{sec:usecase}. In Fig. \ref{fig:stakeholders}, various stakeholders and corresponding questions for which they need an answer, are listed. %Major stakeholders in XAI systems can be summarized as in the following section.

\subsection{Explainability in the Context of Engineers and Scientists} 

 Essentially the mathematical explanation of any ML model requires discussion about the following pivots: 
\begin{itemize}
    \item {\em Objective}: It defines the goals and purpose of the model, and this definition is adequate enough to define what the ML model is trying to achieve.
    \item {\em Scope}: Scope of the ML model is explained clearly and precise enough so that it can answer questions such as, is the scope too small and have limited application, or it is too large resulting in complex and resource-intensive ML model, that is to be used in numerous applications.
\item {\em Model framework}: Description of the system architecture and entire operational scheme is given. This is the description of variables or objectives, and also includes operational details of the model such as governing rules, regulations, and underlying assumptions necessary for system operations.  

\item {\em Modelling and mathematical equations}: 
Explanation includes self-explanatory mathematical equations representing the system structure. These equations define the relationship among various variables, constants, and constraints. In essence, mathematical modeling starts with the general model equation describing inputs to the model and potential output, followed by the detailed architecture of various sub-networks and logic components. These model descriptions also give insight to the intermediate outputs, as well as combinations of these outputs leading to final output of the model. Another important parameter in model explanation is the algorithmic transparency, and it is defined as {\it the ability of the user to understand the model and how a specific output is obtained for certain set of inputs.} 

Moreover, the detailed description of training data is primarily important to further support the transparency of a model and prediction decisions. In this regard, linguistic rules are helpful to interpret training datasets that include descriptions such as, if $x_1$ is high then $y$ is high, and if $x_1$ is low and $x_2$ is high, then $y$ is low (as shown in Fig. \ref{fig:Model}). This description of training datasets make model simulate-able as well as the rule specifications improve interpretability. Feature relevance is another important parameter that contributes to the final output decision and model transparency. For instance, if 5 features $(x1,x2,x3,x4,x5)$ are selected for any model description and $x1$ has prime importance compared to the rest of the four, then it will have for sure, major stake in the prediction decision. Furthermore, one can easily understand that slight change in $x1$ can drastically change the model output.
\end{itemize}

We further elaborate these pivots in Fig. \ref{fig:Model}. Explanation of a model starts with the objective and scope of the model, and `why' we need that model. It is also very important to know about the governing principles of the model along with available data and underlying assumptions (on the unknown parameters). Moreover, the model framework including the system equations is the heart of any mathematically-focused XAI model. The framework captures details about variables, possible parameters, and operational details. Similarly, model equations provide details about the relationship between input and output of a system. Additionally, algorithmic description along with the details about training and test data contribute to the essential part of XAI model. For instance,  algorithmically transparent models are  explorable with mathematical analysis and methods.  Linear models are considered algorithmically transparent because it can be easily understood how these models will behave in particular situations. Additionally, their error surface is known as well. On the other hand,  models with deep architectures are difficult to understand, and their behaviour has to be approximated. Therefore, algorithmic transparency and details of datasets can help in the simulation and (re)testing of the model. Furthermore, feature relevance description clarifies the inner functioning of a model because the relevance score of each variable quantifies the affect of any feature upon the decision making of the model. It also quantifies the importance given to any variable for obtaining the output of a model.

 %\subsubsection{Explainability Discussion} 
\subsection{Explainability of Currently Used ML Models} 

Here, we discuss the explainability of commonly used ML algorithms such as Decision Tree (DT), K-Nearest Neighbors (KNN), and Bayesian models, which are considered as transparent algorithms. Decision trees are hierarchical structures and used for decision making in regression as well as classification problems. A small DT can be easily simulated and managed by a user because the number of features along with their meanings are easily understandable. Therefore, a user can easily simulate DT model and can obtain the
prediction of DT, without the need of any mathematical background. Furthermore, generic rules defining the model and constructing the tree do not change with a change in the dataset, and the model readability is preserved. However, an increase in the size of DT impedes
its full evaluation (simulation) by a human, and therefore, it is considered to be a decomposable model. On the other hand, Bayesian models are probabilistic models and are built in the form of directed acyclic graphical model such that the links between various sets of variables are connected through conditional dependencies. In Bayesian models, relationships between features and the target (connections linking variables to each other) are very clear. For instance, if a Bayesian network is used for the diagnosis of diseases, the probabilistic relationships between diseases and symptoms is intuitively clear, and probability of presence of any diseases can be easily determined by inspecting the symptoms. However, in the presence of overly complex and cumbersome variables, statistical relationships cannot be interpreted easily, and can only be analyzed using mathematical
tools. In case of KNNs, complexity of the model and detailed description of number of
variables, and similarity measures are pretty intuitive for humans and, KNNs are easily simulate-able. Therefore, the afore-mentioned models are considered as transparent models. These are also decomposible as various variables can be decomposed and analyzed easily. However, when the similarity measure of variables is very high and too complex to decompose, then mathematical tools are used for explanation. 

\section{Related Work}
Here we discuss the existing literature on the general XAI techniques and their use in different domains including IoT. %In the later sections, we provide detailed discussion on the applications of XAI in IoT networks.

In \cite{Prashan}, the authors presented explainable causal models that augment the underlying AI models. They derived their model by analyzing 398 explanation dialogues across six different dialogues (using grounded theory). Their proposed model accurately defines the structure and the sequence of an explanation dialog, and appears to support natural interaction among human audience compared to explanations obtained from the existing models.
In the same spirit, the authors in \cite{Katharina}  examined the impact of virtual agents in XAI on the trustworthiness of autonomous systems perceived by human users. They conducted a user study based on a simple speech recognition task and found significant evidences that the integration of virtual agents and XAI leads to an increased trust in the autonomous intelligent system. Their results also showed that users trust AI system which use visual agents more than AI system interpreting mere explanation. 

In \cite{Wolf}, the authors proposed a scenario-based design for system development and leveraged various possible scenarios to anticipate and predict new scenarios. The authors  presented a case study of aging-in-place monitoring, and demonstrated the relevance of the scenario-based design to XAI design practice. In \cite{Arrieta}, the authors discussed  recent contributions related to the explainability of different ML models and related challenges. They also presented the concept of responsible AI, and related guidelines for explainable and responsible AI such as, fairness, accountability and privacy. In \cite{Liao}, the authors identified the gaps between the current XAI algorithmic practices and to create related explainable AI products. They created the details of an XAI design space by developing algorithm informed XAI question bank. This question bank can be used as a baseline to create user-centered XAI and incorporate important explainability features into the XAI systems.     

In \cite{Quanshi}, the authors presented a generic interpretable model to learn interpretable convolutional filters used in the construction of interpretable CNN. These filters learn to represent the same object portion across all the images fed to CNN, and do not require any additional object annotations or textures (just training data). During the learning process, the interpretable filters are assigned automatically to each convolutional layer. These filters represent explicit knowledge of various parameters in each of the CNN's convolutional layers, such as, encoded logic of CNN, what patterns are extracted from input image, and how prediction is made, etc. 
%\section{Stakeholders in XAI and Requirements}

%\subsubsection{XAI and autonomous car: use-case from computation and communication perspectives}
\section{Use-case: Autonomous Car}
\label{sec:usecase}
To discuss the explainability from an engineering perspective, we take the autonomous car use-case that extensively uses artificial intelligence \cite{Hussain2018,Ma2020}. We discuss explanatory requirements in autonomous cars as a use-case of XAI and its perspective of impacting human lives. In essence, we consider the key components of autonomous car such as perception, object detection, and actuation and discuss the necessity, role, and techniques used to realize and incorporate XAI.

\subsection{General Objective, and Transparency and Explainability Requirement for the XAI Model}

The general objective of an XAI model for any autonomous vehicle is relatively straightforward, i.e., safe operation (perception, planning, and actuation) of an autonomous vehicle. Also, it should incorporate the rendering of surrounding environment in the navigational decision making. A system should be intelligent enough not only to predict the possible changes in the environment but should also be adaptable to those changes. In this regard, reinforcement learning techniques can be used to interact and learn from the environmental changes. To date, reinforcement learning has been used for different functionalities in autonomous car such as navigation \cite{Isele2018}, planning \cite{YOU2019}, and control \cite{Xia2016}.  Also, for autonomous cars, the scope should be clearly defined such as, is vehicle designed to operate in north American traffic or is it equally capable of operating globally? In other words, the context is important.

It is essential to discuss the need for transparency and explanation of the AI-related solutions in autonomous car. If an autonomous car makes an unexpected decision (such as wrong direction, wrong turn, sudden brakes, problems with object identification, crashing into other objects, or fail to apply brakes, etc.), it is essentially important to understand why it happened, for a couple of reasons. The first reason is to fix the problem and to improve the user experience and trust in the autonomous car technology. The second reason is that in case of an accident, it is critically important to perform forensics and figure out the source of the accident. That will be possible only if the decisions taken by the autonomous car are transparent and explainable. %on the actions before accident are transparent. 
From this perspective, XAI is about the improvement, transparency, explanability, and optimization of the decisions taken by autonomous cars. % as well as the transparency of the decisions. 
 
Shen et al. \cite{shen2020explain} thoroughly investigated when an explanation is needed during autonomous car drive and how does the explanation changes with context? This study reveals that the explanation purely depends on the driving scenarios and a real-time text explanation can be provided to the passengers based on the analysis of the data obtained from on-board camera. The study reveals a number of important aspects of the need for explanation in autonomous cars. For instance, explanation is crucial for near-crash scenarios and un-expected decrease in speed. While some scenarios call for decrease in speed such as stop sign, while others do not. The explanation can be tightly coupled with the passengers' expectations. For instance, if the speed was decreased and not expected by the passenger, then explanation is provided to the passenger.  
 
There is an ethical perspective of appropriate responses of a car to certain situations, and ethicists are curious to know the decision variables. However, automakers (engineers) need explanations about how and what an autonomous car will do when an accident is inevitable? Will it prioritize the protection of the occupants or pedestrians ({\it the trolley problem})? Will it apply emergency breaks or crash into an object? Engineers and scientists can track the data-set (with various driving scenarios) by explaining how the decision was made, and thus make it easier to debug, improve the decision procedure, test, and so on.  %for them to understand, assess and improve the decisions made by the car. 
Similarly, explanation about route selection is also important for algorithmic understanding. More precisely, the question of whether a good route is the shortest recommended route, or the car should opt for a longer route but scenic, can be taken into account. Additionally, passengers also need explanation (in general) about how ethical decisions are made by the car. As the occupants have to decide, if they are comfortable traveling in a vehicle, that is designed to make specific decisions in special circumstances. %Furthermore, 

\subsection{Mathematical Perspective of the XAI Model in Autonomous Cars}

\begin{figure}
%\centerline{\includegraphics[scale=0.6]{Model.png}}
% [width=8in,height=6in]
\centerline{\includegraphics[width=3.5in]{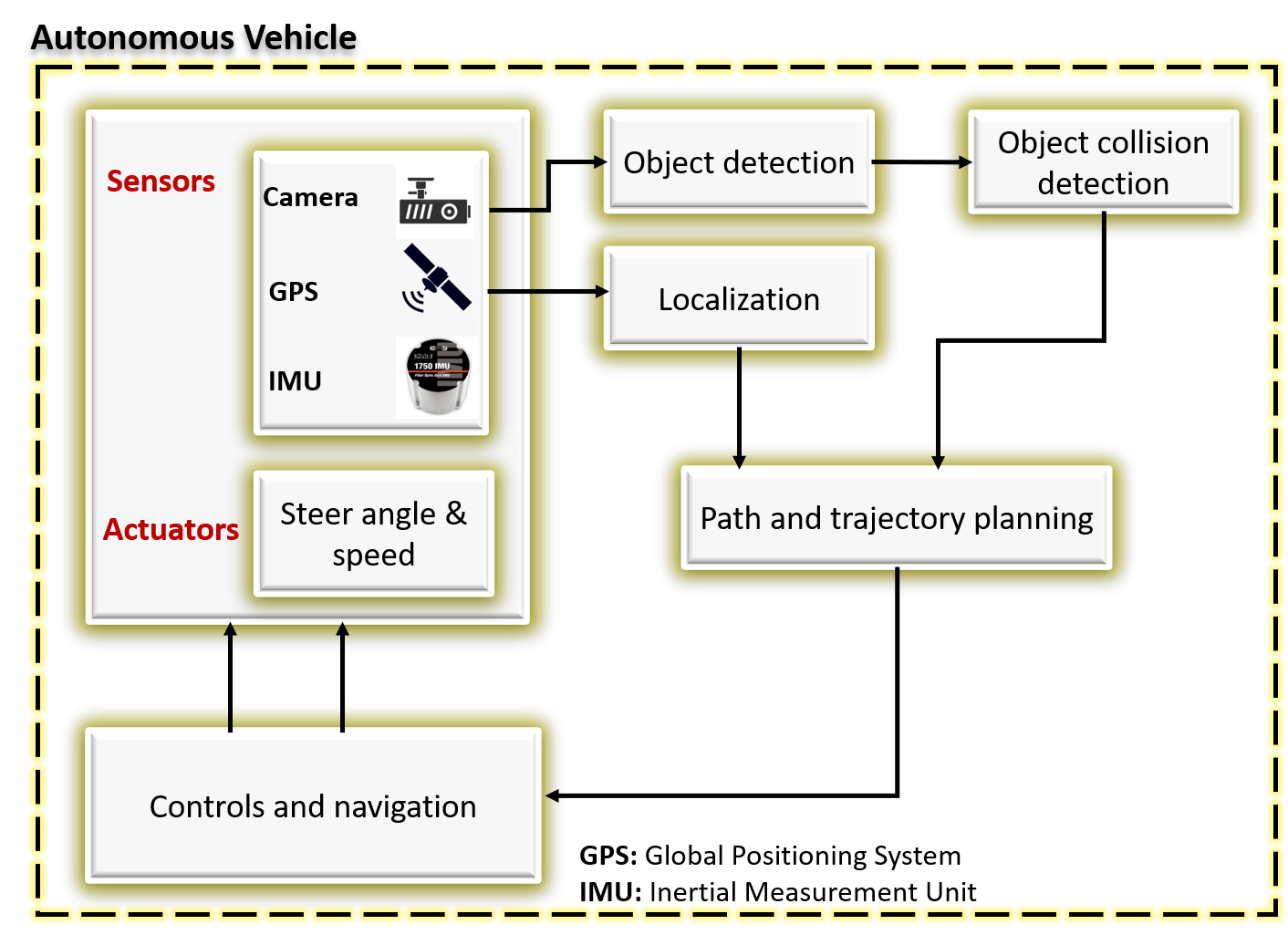}}
\caption{Autonomous Vehicle Decision Model}
\label{fig:AV}
\end{figure}

We dive into the mathematical perspective of an XAI model for autonomous cars. We note that it is a complex system consisting of many interacting agents including cars (including software and hardware), road semantics, pedestrians, and car position and controls. Interactions among these agents can be modeled through differential equations, and these equations are used to understand the behaviour of any autonomous model. Following the foundational pivots defined in Fig. \ref{fig:Model},  XAI model framework (mathematically  driven) of an autonomous car is shown in Fig. \ref{fig:AV}. 
In the following, we discuss the types and levels of explanation required of different functional entities in autonomous cars.

\subsubsection{Object detection and classification} 
What are various types of external and internal sensors such as LiDAR, Radars, cameras, or Internet of Things (IoT) devices used in the image sensing? How data fusion is performed on data obtained from different sources? Details of pattern recognition algorithms used in image classification (after images obtained through sensors) are required. Since various types of images (environmental) data is received through Advanced Driver Assistance Systems (ADAS) data, reduction algorithms are used for filtering the irrelevant images. Details of the type of data reduction algorithm should be provided, i.e., is it  Principle Component Analysis (PCA), Histograms of Oriented Gradients (HOG), or Support Vector Machines (SVM)? 

After image classification, object detection is performed which leads to the following questions. Which Regression algorithm is used for object detection? What are the dependent and independent variables used in the regression algorithm? It is worth mentioning that latent structures are exploited to interpret the end-to-end learning and inference models in object detection \cite{Wu2019ICCV}. The authors in \cite{Wu2019ICCV} use two-stage convolutional networks based on regions in an image and generate rationale which serves as explanation. The authors combined top-down grammar models with the bottom-up convoluntional networks. They used hierarchical representation of the Region of Interest (RoI) instead of widely used flat structure and also enriching the RoIs for interpretable object detection. Similarly, Soares et al. \cite{Soares2019} use 0-order fuzzy rules to provide human understandable 'IF-THEN' representation pertaining to different decisions taken by autonomous cars (for instance acceleration, lane change, etc.). The proposed classifier not only improves the classification accuracy but also focuses on the density of the features that contribute to the underlying decision by leveraging massively parallel 0-order fuzzy rules.  

\subsubsection{Object localization, perception, and movement prediction} 
Regression algorithms are used for motion prediction and to create a statistical relational model of the detected object’s current and future position. It leads to questions such as, which regression algorithm is used for position prediction and why? Decision Forest regression, neural network regression or Bayesian regression? Is regression algorithm used for short prediction or long time global prediction? Which localization algorithms are used and what are the reasons for selecting particular algorithms? To the best of our knowledge, the explanation has not been introduced to the object localization and movement prediction of autonomous cars. However, efforts have been put forth in introducing interpretation to perception in autonomous cars. 

Here we discuss the existing work in this direction. In \cite{Kim2017ICCV}, Kim et al. used visual attention networks for the interpretation of perception developed by the autonomous cars. Visual attention networks \cite{Xu2015} play pivotal role in interpretable learning where these networks provide visual attentions to areas in an image that are influenced by the network and the areas attended by the underlying networks. These networks are inherently suitable for autonomous cars use-case where real-time images are captured during driving. In this work, Kim et al. applied post-processing to the output of the visual attention network on the images from the autonomous car and then determined the effect of the attention areas on the end-to-end output of the network. Then the areas of image that have causal effect on the end-to-end network learning are identified and presented to the user. In other words, the authors detected the visual saliency in images. Furthermore, this work also concluded that the visual attention heat maps are essentially important for suitable explanation and do not degrade the efficiency of object detection and perception. Additionally, Kim et al. also reported that causal filtering improves the explanation by removing unnecessary features that do not induce the output. 

In the realm of perception, visual intelligence plays a pivotal role in autonomous cars and the explainability related to visual intelligence is important at par. In this regard, Zhang et al. \cite{Zhang2019CVPRW} claim that the existing methods of visual perceptions have inherent disadvantages in real-world scenarios such as extreme weather, low illumination, etc. and these might be caused due to biasedness in the training dataset. Although interpretability of learned representations by the deep networks partially addresses the afore-mentioned challenge; however, external factors related to test data are equally important. Zhang et al. proposed an explainable AI evaluation where they tried to include human domain knowledge to the variances in test data that would augment the semantic concepts and the output of the underlying algorithms used for visual perception. They used Ridge Regression  to find deterministic relationships and then test the output of a specific algorithm.

%by considering variance in its output in regards to the variance in test data. After that, human domain knowledge is mapped to the disentangled variables to interpret their relationship to the output.

Similarly, in perception, visual regions and their significance as well as importance are of supreme importance. Therefore, it is essential to focus on such regions in visual scenes while performing perception in autonomous driving. Yang et al. \cite{Yang2019} proposed a framework to interpret the role of CNN in end-to-end learning in understanding the driving scenarios. The proposed framework determines the regions in visual scenes that have major contributions in the perception. The authors ranked the importance of the regions in scenes to determine how different scenarios and the regions in scene affect the performance of CNN in perception.  

Motion planning is the next step to perception and prediction in autonomous driving. The output of the perception and prediction (that predicts the future positions and trajectories of the objects detected and perceived) is used for motion planning in terms of safe trajectory. However, this is not an easy task as it may result in sub-optimal performance at individual tasks (such as detection, perception, and prediction). In other words, Zeng et al. \cite{Zeng2019} pointed out that optimization in one local task does not universally translate to other tasks. In this regard, Zeng et al. \cite{Zeng2019} proposed a new end-to-end motion planning framework in autonomous cars that is interpretable. The proposed scheme takes LiDAR data as inputs and interprets the intermediate representation of the input data into future trajectories. It also identifies the degree of goodness of each possible position in the planned trajectory that the autonomous car can take. This work is extended to \cite{sadat2020perceive} to include interpretable perception and prediction as well as motion planning. Furthermore, in the extended work, the same authors implemented learning from human demonstration to imitate the real driving scenarios.

\subsubsection{ Action decision} 
Decision is the pinnacle to the autonomous car operation and as a result of perception, a decision is taken by the autonomous car. This decision could be the result of a learning process which warrants explanation as we mentioned before. To this end, Xu et al. \cite{Xu2020} proposed an enhanced pipelined system based on action-inducing objects. More precisely, the objects that are directly related to the driving tasks. For example, pedestrians walking on a side-walk are not as important as pedestrian crossing the roads. The latter are referred to as action-inducing objects. Xu et al. proposed an architecture where every action-inducing object is responsible for an action (for instance, an object detected on the road might induce the autonomous car to apply brakes) provided with an explanation and they argued that the set of explanations is finite in a given scenario. To achieve this goal, the authors annotated a dataset for different commands pertaining to driving and their respective explanations. Furthermore, the authors solved the problem of object reasoning keeping in mind the scene context with a goal to detect objects that induce the decision, i.e., objects that may create hazard in driving. And the detection includes the respective explanation about the prediction of action taken by the autonomous car.

\subsubsection{Navigation and control} 
Navigation and control are important at par, to realize autonomous driving. Control algorithms are responsible for the action decisions in autonomous driving, therefore, the explanability and transparency of these algorithms is essential. In terms of transparency and explanability, some important questions should be answered. For instance, what type of decision matrix algorithm is employed for making specific decisions in specific scenarios? The question whether a car needs to apply brake or take a turn, is directly related to the level of confidence in the recognition, classification algorithms as well as on the prediction of the next movement of detected objects. How many decision models contribute towards this decision, and how these models are trained? How decisions made by individual models are combined to make the overall prediction while considering the possibility of decreasing errors while making decisions?

Another very important description (explanation) is related to motion dynamics of a vehicle considering elements such as motor dynamics, brake system dynamics, and roll and pitch motion, etc. Is six Degree of Freedom (DOF) model used (considering all possible movements) or a simplified kinematic model is used (that does not take into account dynamics such as roll, pitch and Z motion)? What are various set of equations that describe the vehicle motion in the X-Y plane? Do these system equations describe the drive-train dynamics, brake dynamics, wheels dynamics and the engine dynamics clearly?
Furthermore, car position at every instant has to be updated and depends on its current velocity, acceleration, and deceleration due to other vehicles and random events. Is such deceleration due to extra fuel consumption or due to some road blocker? What are the deciding factors for this deceleration? To this end, according to the best of our knowledge, not much work has been done in this direction. There exist only a few works that look into the explanabitlity and transparency in navigation. For instance, He et al. \cite{he2020} proposed an explainable DRL mechanism for navigation in autonomous cars and autonomous drones. The proposed scheme aims at both non-expert and expert users of the system to provide them with explanation on the decision taken by the model. These explanations are either visual or textual. In essence, the proposed scheme used post-hoc explanation for the trained navigation policy.

From the preceding discussion, we can conclude that there are still many questions that are unanswered and warrant further investigation in autonomous cars from the perspective of XAI.

%\section{NONAME section}

%\subsection{Explainablity Analysis of various ML algorithms }We use the classification of transparent models( made in the above lines)to describe level of transparency of various ML techniques. 
%\subsubsection{Transparent ML Techniques}
%\subsubsection{Opaque  ML Techniques}\cite{Tabrez}\cite{Alexander}\cite{Thilo}\cite{Tongyu}\cite{Rahul}
%\subsection{XAI and various IoT Applications }\cite{TED}, (((In \cite{Brian}, Expert Systems and Bayesian Network, recommender systems and NLG are discussed.))) \cite{Nicolota}

%Authors in \cite{Amina}, presented framework for explanations of recommendations made by  a recommender system. Authors used knowledge extraction method for expalinability of the recomender systems recommendations. This approach is based on modeling the knowledge about users and user's selected items, followed by the extraction of explanations by defining certain rules. 
%\section{Future Challenges} \textcolor{red}{possibly Rasheed }
%Various challenges are associated with providing meaning full explanations \cite{Ted}
%\subsection{Adversarial AI (should we keep it ?)} 
%\subsection{Standardization}
%\subsection{Social challenges }
%\subsection{Business Challenges}

\section{Conclusion and Open Challenges} %{\fatima STILL IN PROGRESS}

%'To whom the explanation of a model is needed?' is one of the most important questions in the context of XAI. On the other hand,
The past few years have seen extensive interest in the field of XAI. Due to the varying stakeholders, the same XAI can be approached differently. In this paper,% we looked at XAI from an Engineering perspective and looked at various stakeholders. 
 we have shed light on the need for XAI, how it works generically, and how can we look at XAI from an engineering perspective. Then we have taken the autonomous car as a use-case and discussed different components of autonomous driving from explainability and transparency standpoint. We have reviewed the existing works that focus on these different aspects of  autonomous cars and provide transparent, interpretable, and explainable solutions. %The most important question in context of XAI is; to whom is the model explainable and interpretable not the inquiry about, is the model interpretable or not. To be continued......

Despite the surge in XAI solutions in the literature, there are still open challenges that need attention from the research community. In the following, we discuss the open challenges in XAI.

\subsubsection{Generalization of XAI} 
From the preceding discussion, we note that XAI, at least at the moment, is highly environment- and domain-dependent. Generalization of XAI might take long time to realize; however, the need for XAI cannot be ignored, even more so with the realization of GDPR. The real question is `when' and not `why'. However, further investigation is needed in this direction. More precisely, due to the variation in stakeholders and their requirements, it will be challenging to generalize XAI; however, one possible solution would be to focus on domain-level explanation and work towards local generalization from domain standpoint.

\subsubsection{Adaptation of XAI}
%We also note that 
There have been noteworthy efforts in the field of XAI covering different gaps in different domains, but there is still a struggle with the applicability of XAI. Among other reasons, the main reason is the variety and level of explanations needed for different stakeholders. For instance, in case of an autonomous car, the same AI model might need different level of explanation for end-users, developers, designers, and so on. The adaptation of XAI is, by far, one of the biggest challenges that needs further research. Domain adaptation might be a good starting point to make the XAI applicable as generally as possible.

%Secondly, an XAI model in one scenario may perform differently in another scenario with different data, therefore, efforts towards generalization of XAI model is needed. 
\subsubsection{Security of XAI}
The adversarial learning is gaining ground and its effects on XAI are still under-explored. The effect of adversarial machine learning is unquestionably an important topic of research, covering both attacks on ML and DL models, and the use of ML and DL models in adversarial settings \cite{Papernot2018,WANG2019}. The adversarial ML and DL models (or the use of ML and DL in adversarial settings) in XAI will be an interesting topic to investigate. Perturbations in input data to learning models, bias, and fairness are the key enablers for the security of AI models. These features will be  of paramount importance in XAI and must be investigated.  

\subsubsection{XAI and Responsible AI}
Although Responsible AI (RAI) embodies the explainability, interpretability, and transparency, these components must be put into perspective. In other words, the aforementioned features of AI are currently researched under the umbrella of XAI. But their real-world applications should be investigated as RAI where the effect of XAI on usability of AI models is subject to further investigation. For instance, among the research community, the rationale for XAI is still controversial. Furthermore, apart from technical perspective, the social and legal dimensions are also important at par, to realize the XAI in real-world scenario, which is what RAI covers. Therefore, there are a lot of research opportunities in this direction.   

\subsubsection{Performance of XAI}
Although XAI adds the necessary features to the baseline AI models, it is also very important to assess the performance of AI models with the additional features related to explanation. It is imperative to think that the new features will likely incur overhead and may affect the accuracy. To date, some works  in the literature have focused on the performance of XAI \cite{das2020,Tritscher2020,lin2020}. Further investigation is needed in this direction. Also, the effect of XAI on the original model (whether good or bad) also warrants further in-depth investigation to open new avenues of research.

\bibliographystyle{IEEEtran}
\bibliography{main}
\end{document}